\documentclass[11pt]{article}

\usepackage{graphicx}
\usepackage{amssymb,amsmath}
\usepackage{color}
\usepackage{natbib}

\newenvironment{itemize*}{
\begin{itemize}
\setlength{\parskip}{0em}
\setlength{\topskip}{0em}
}
{\end{itemize}}

\newenvironment{enumerate*}{
\begin{enumerate}
\setlength{\parskip}{0em}
\setlength{\topparskip}{0em}
}
{\end{enumerate}}

\definecolor{mygray}{gray}{.30}

\newtheorem{prop}{Proposition}

\newcommand{\beq}{\begin{equation}}
\newcommand{\eeq}{\end{equation}}
\newcommand{\beqa}{\begin{eqnarray}}
\newcommand{\eeqa}{\end{eqnarray}}
\newcommand{\beqas}{\begin{eqnarray*}}
\newcommand{\eeqas}{\end{eqnarray*}}

\newcommand{\bit}{\begin{itemize}}
\newcommand{\eit}{\end{itemize}}
\newcommand{\bits}{\begin{itemize*}}
\newcommand{\eits}{\end{itemize*}}
\newcommand{\benum}{\begin{enumerate}}
\newcommand{\eenum}{\end{enumerate}}
\newcommand{\benums}{\begin{enumerate*}}
\newcommand{\eenums}{\end{enumerate*}}
\newcommand{\comment}[1]{}

\newcommand{\mmp}[1]{\textcolor{blue}{\em MMP: {#1}}}
\newcommand{\techreport}[1]{}

\newcommand{\mf}[1]{}

\newcommand{\mydef}[1]{{\em #1}}

\newcommand{\rrr}{{\mathbb R}}

\newcommand{\clust}{{\cal C}}

\newcommand{\loss}{\ensuremath{\operatorname{Loss}}}

\newcommand{\dataset}{{\cal D}}

\renewcommand{\mmp}[1]{}

\newcommand{\lca}{\operatorname{lca}}

\newcommand{\xijt}{x_{ij}^t}

\newcommand{\yijt}{y_{ij}^t}
\newcommand{\yijtunu}{y_{ij}^{t+1}}

\newtheorem{theorem}{Theorem}

\setcitestyle{authoryear, open={(},close={)}}

\title{Guarantees for Hierarchical Clustering by the Sublevel Set method}

\author{Marina Meil\u{a}}
\date{October 16, 2019}

\begin{document}
\maketitle

\begin{abstract}
\citet{meila2018tell} introduces an optimization based method called the Sublevel Set method, to guarantee that a clustering is nearly optimal and ``approximately correct'' without relying on any assumptions about the distribution that generated the data. This paper extends the Sublevel Set method to the cost-based hierarchical clustering paradigm proposed by \cite{Dasgupta:16}.
\end{abstract}

\section{Introduction}
Compared to (simple) clustering data into $K$ clusters, hierarchical
clustering is much more complex and much less understood. One of the
few seminal advances in hierarchical clusterings is the
introduction by \cite{Dasgupta:16} of a general yet simple paradigm of
hierarchical clustering as loss minimization. This paradigm was
expanded by \cite{charikar:16} and \cite{royPokutta:hierarchical16}. The latter work also
  introduces a new set of techniques for obtaining hierarchical
  clusterings by showing that optimizing the loss can be relaxed to a
  Linear Program (LP).

This paper introduces the first method to obtain optimality guarantees in the context of hierarchical clustering. Specifically, it is shown that the {\em Sublevel Set (SS)} paradigm invented by \cite{meila2018tell} for simple, non-hiearchical clustering, can be extended as well to hierarchical clustering. The main contribution is show that there is a natural distance between hierarchical clusterings whose properties can be exploited in the setting of the SS problem we will present in Section \ref{sec:ss}. 

The Sublevel Set method produces {\em stability theorems} of the following form.
\begin{theorem}[Informal Stability Theorem] \label{thm:stabi} If a clustering $\clust$ has low enough loss for a data set $\dataset$, then, subject to some conditions verifiable from the data, any other clustering $\clust'$ that has lower or equql loss to $\clust$ cannot
be more than $\epsilon$ different from $\clust$.
\end{theorem}
When a stability theorem holds in practice, it means that $\clust$ is not just a ``good'' clustering; $\clust$ must be the only good clustering
 supported by the data $\dataset$, up to small variations. This
property is called \mydef{stability}. It is obvious that, even though a Stability Theorem does not guarantee optimality, it implies that the optimal clustering of the data is within distance $\epsilon$ of $\clust$. The value $\epsilon$ which bounds the amount of variaation in the above theorem, defines a ball of radius $\epsilon$ around $\clust$ that contains all the good clusterings, including the optimal one. This ball is called an {\em optimality interval (OI)} and with a slight abuse we will also refer to its radius $\epsilon$ as an OI.

The main result of this paper is Theorem \ref{thm:ss} in Section
\ref{sec:ss} which will give an OI for hierarchical clustering in the
paradigm of \cite{Dasgupta:16}, along with a simple algorithm for
calculating the OI $\epsilon$, based on the LP relaxation of
\cite{royPokutta:hierarchical16}. We formally define the distance in
the space of hierarchical clusterings in which the OI is to be
measured in the next section.

\section{Preliminaries and a distance between hierarchical clusterings}
\label{sec:background}
\paragraph{A loss function for hierarchical clustering}
Let $n$ be the number of points to be clustered, and $T$ be a hierarchical clustering, or {\em tree} for short,  whose leaves are the $n$ nodes. All trees have $n$ levels, and between one level and the level below, a single cluster is split into two non-empty sets; at level $t$, with $t=0:n-1$, there are $n-t$ clusters. A tree $T$ can be represented as a set of $n$ matrices $X(T)\equiv X=[x_{ij}^t]_{i,j=1:n}^{t=1:n-1}$. The variable $x_{ij}^T=1$ if nodes $i,j$ are separated at level $t$ of $T$, and 0 otherwise. The levels of the tree are numbered from the botton up, with 0 the level of all leaves ($x_{ij}^0\equiv 1$ implicitly), and the highest split at level $n-1$. Let $X^t=[\xijt]_{i,j=1:n}$ denote the matrix representing the clustering at level $t$. Each $X^t$ matrix is symmetric with 0 on the diagonal. Note also that $\sum_{t=1}^{n-1} \xijt = l_T(i,j)$, where $l_T$ is the path length from $i$ or $j$ to their {\em lowest common ancestor} ($\lca$). 

Denote by $S\in\rrr^{n\times n}$ a symmetric matrix of similarities, such that $S_{ij}$ is the cost of {\em not} having $i,j$ together at any level in the clustering. The cost of a hierarchical clustering $T$ is the sum of the costs for each pair of nodes $i,j=1:n$, and each level $t$. It is assumed that $S_{ii}\equiv 0$ to simplify the algebraic expressions.  This cost was introduced by \cite{Dasgupta:16} who showed that it has many interesting properties. 
With the $X$ notation for a hierarchical clustering, the cost can be re-written as shown by \cite{royPokutta:hierarchical16}
\beq\label{eq:loss}
\loss(S,X)=\sum_{i,j=1}^n\sum_{t=1}^{n-1} S_{ij}\xijt+\sum_{i,j=1}^nS_{ij}\;=\;\sum_{i,j=1}^nS_{ij}\left(\sum_{t=1}^{n-1}\xijt\right)+\sum_{i,j=1}^nS_{ij}.
\eeq
Note that the second term is a constant independent of the structure of $T$. 
In \cite{royPokutta:hierarchical16} is it shown that minimizing $\loss$ over hierarchical clusterings $X$ can be formulated as an Integer Linear Program, which can be relaxed as usual to a Linear Program (LP).

\paragraph{The Matrix Hamming distance between hierarchical clusterings}
For any two matrices $A,B\in\rrr^{n\times n}$, let $\langle A,B\rangle=\sum_{i,j=1}^na_{ij}b_{ij}$ denote the {\em Frobenius scalar product}, and $||A||^2_F=\langle A,A\rangle$ denote the {\em Frobenius norm}, squared. For two hierarchical clusterings $X,Y$, we define   $\langle X,Y\rangle=\sum_{i,j=1}^n\sum_{t=1}^{n-1}\xijt y_{ij}^t=\sum_{t=1}^{n-1}\langle X^t,Y^t\rangle$. This is clearly a scalar product on the space of hierarchical clusterings, and
\beq
||X||^2_F\;=\;\langle X,X\rangle=\sum_{i,j=1}^n\sum_{t=1}^{n-1}(\xijt)^2\;=\;\sum_{i,j=1}^n\sum_{t=1}^{n-1}\xijt.
\eeq
The last equality holds because all $\xijt$ are either 0 or 1.
Moreover, in \cite{Dasgupta:16} it is proved that 
\beq
\sum_{i,j=1}^n\sum_{t=1}^{n-1}\xijt\;=\; \tfrac{n^3-n}{3}.
\eeq
Therefore, we have the following simple results.
\begin{prop} \label{prop:xnorm} For any hierarchical clustering over $n$ points $X$, $\|X\|^2_F=\tfrac{n^3-n}{3}$.
\end{prop}
For any two binary matrices $A,B\in\{0,1\}^{n\times n}$, we define the {\em Matrix Hamming (MH) distance} $d_H(A,B)$ to be the number of entries in which $A$ differs from $B$, i.e. $d_H(A,B)=\sum_{i,j=1}^nA_{ij}\oplus B_{ij}$, where $\oplus$ denotes the exclusive-or Boolean operator. We further extend the MH distance to hierarchical clusterings by $d_H(X,Y)=\sum_{t=1}^{n-1}d_H(X^t,Y^t)$. 
\begin{prop}\label{prop:dham} Let $X,Y$ be two hierarchical clusterings over $n$ points. Then $d_H(X,Y)=\|X-Y\|^2_F$.
\end{prop}
{\bf Proof} 
\beqa
||X-Y||^2_F&=&||X||_F^2+||Y||_F^2-2\langle X,Y\rangle\\
&=&2\tfrac{n^3-n}{3}-2\langle X,Y\rangle\\
&=&\sum_{i,j=1}^n\sum_{t=1}^{n-1}\left(\xijt+y_{ij}^t-2\xijt y_{ij}^t\right)\\
&=&\sum_{i,j=1}^n\sum_{t=1}^{n-1}\left(\xijt \oplus y_{ij}^t\right)\\
&=&d_H(X,Y).\label{eq:dham} \hfill\Box
\eeqa
For simple, non-hierachical clusterings, $d_H$ is equivalent to the unadjusted Rand Index \cite{M:compare-jmva}. The (unadjusted) Rand Index has long been abandoned in the analysis of simple clusterings because, when the number of clusters $K$ is larger than 4 or 5, all ``usual'' clusterings appear very close under this distance. This was further formalized by \cite{M:compare-icml05}. 

This disadvantage for simple clusterings may turn to be an advantage in the hierarchical setting. We expect that, for small values of the level $t$, near the leaves of the tree, $d_H$ will be very small w.r.t. the upper bound $n^2-n$. Indeed, for $t=1$, $X^1$ and $Y^1$ contain each one pair of merged points, hence $d_H(X^1,Y^1)=2$, whenever $X^1\neq Y^1$. Hence, in $d_H(X,Y)$, the levels of the cluster tree below the very top ones are strongly down-weighted, letting the top splits dominate the distance.

\section{Sublevel Set method}
\label{sec:ss}
Now we are ready to apply the SS method of \cite{meila2018tell} to a hierarchical clustering. 

From Proposition \ref{prop:dham} it follows that maximizing $d_H(X,Y)$ is equivalent to minimizing $\langle X,Y\rangle$. Therefore, we can obtain a stability theorem and an OI  as folows. Assume we have data $S$, a hierarchical clustering $T$, obtained by minimizing $\loss(S,X)$ as well as possible. Hence we assume $X$ is fixed;  $Y$ is any other arbitrary other clustering. 
We define the following optimization problem, which we call a {\em Sublevel Set} problem. 
\beqa
(SS)\quad \delta\; =&\min_{\,Y}&\langle X,Y\rangle\\
&\text{s.t.}&\loss( S,Y)\leq \loss( S,X) \label{eq:ss-sublevel}\\
&& \yijt\geq \yijtunu,\text{ for all $t,i,j$}\label{eq:ss1}\\
&&\yijt+y_{jk}^t\geq y_{ik}^t,\text{ for all $t,i,j, k$}\label{eq:ss2}\\
&&\sum_{j\in S}\xijt \geq |S|-t,\text{ for } t,\, S\subseteq [n] \text{ with } |S|>t,\,i\in S \label{eq:ss-powerset}\\
&&\xijt\in[0,1],\text{ for all $t,i,j$}\label{eq:ss3}\\
&&\xijt=x_{ji}^t,\,x_{ii}^t=0,\text{ for all $t,i,j$}\label{eq:ss4}
\eeqa
The problem above maximizes $\|Y-X\|^2_F$ over a relaxed space of non-binary matrices $Y$ that satisfy contraints \eqref{eq:ss1}--\eqref{eq:ss4}; all matrices representing cluster trees also satisfy these constraints. Constraint \eqref{eq:ss-sublevel} restricts the feasible set to those $Y$ that have lower or equal cost to $X$, therefore this set is called a {\em sublevel set} for $\langle S,Y\rangle$. We note that $\loss(S,Y)$ is linear in $Y$, therefore the (SS) problem is a Linear Program.

The Sublevel Set problem above follows the (LP-ultrametric) problem formulation from Section 4 of \cite{royPokutta:hierarchical16}, with the addition of the sublevel set constraint \eqref{eq:ss-sublevel} and replacing $S$ with with $X$ in the objective. Note that in \eqref{eq:ss-powerset} are an exponential number of constraints; \cite{royPokutta:hierarchical16} claim that the LP can still be optimized in poly$(n,\max \log S_{ij})$ operations. In \cite{meila2018tell} the SDPNAL software \cite{yang2015sdpnal} was used, and this software can also solve LPs.

\begin{theorem}[Stability Theorem for hierarchical clustering] \label{thm:ss} Let $S,X,n$ be defined as above, and let $\delta=\delta(S,X)$ be the optimal value of the (SS) problem. Then, any other clustering $Y$ with $\loss(S,Y)\leq \loss(S,X)$ satisfies $d_H(Y,X)\leq \epsilon$, with $\epsilon=2(\tfrac{n^3-n}{3}-\delta)$.
\end{theorem}
The proof of the Theorem is immediate from the constraint
\eqref{eq:ss-sublevel} and Propositions \ref{prop:xnorm} and
\ref{prop:dham}.  In more detail, $Y^*$ the solution of (SS) may not
be an integer solution. However, the (SS) problem guarantees that for
any hierarchical clustering $Y$ that has $\loss(S,Y)\leq \loss(S,X)$,
$\langle X,Y\rangle \geq\langle X,Y^*\rangle \geq \delta$. From
Propositions \ref{prop:xnorm} and \ref{prop:dham}, it follows that
$d_H(X,Y)\leq \epsilon$. Hence, all good clusterings of the data must
be in a ball of radius $\epsilon$ from $X$, and
$\epsilon=2(\tfrac{n^3-n}{3}-\delta)$ gives an {\em Optimality
  Interval} for $T$, in terms of Hamming distance.

The (SS) optimization problem can be solved numerically, to obtain $\epsilon$. 
\benum
\item Given similarity matrix $S$, use a hierarchical clustering method to obtain a clustering $X$.
\item Compute $\loss(S,X)$.
\item Set up and solve the (SS) problem by calling an LP solver. Let $\delta,Y^*$ be the optimal value and optimal solution of (SS).
\item Compute $\epsilon=2(\tfrac{n^3-n}{3}-\delta)$
\item[Output] The optimality interval $\epsilon$
\eenum
In \cite{M:equivalence-ML10} are formulas that bound the clustering {\em misclassification error distance} (also known as {\em earthmover's distance}) by the Matrix Hamming distance. They can be used to translate the $\epsilon$ bound into a bound on the more intuitive missclassification distance; this would come with a further relaxation of the bound. 
\comment{What is this??
\beq
\sum_t(\xijt-\yijt)\;=\;d_{(T(X)}(i,j)-d_{(T(Y)}(i,j)
\quad
d_H(X,Y)\;=\;\sum_t d_H([\xijt], [\yijt])
\eeq
}

\section{Conclusion}
\label{sec:discussion}
In this paper we have used the SS method to develop an algorithm that outputs optimality guarantees, in the form Optimality Intervals in the metric space of hierarchical clusterings defined by the matrix Hamming distance $d_H$. Besides guaranteeing (sub)-optimality, the OI also guarantees stability, provided that it is small enough. In other words, when the OI is small, not only is the cost of the estimated clustering $X$ almost optimal, we are also guaranteed that there is no other very different way to partition the data that will give the same or better cost. Much remains still to be studied, in particular how small must $\epsilon$ be for the bound to be truly meaningful.

\section*{Acknowledgement}
The author acknowledges support from NSF DMS award 1810975. This work was initiated at the Simons Institute for Theoretical Computer Science during the time Meila was a long term visitor, and continued at the Institute for Pure and Applied Mathematics (IPAM). The author gratefully acknowledges a Simons Fellowship from IPAM, that made her stay during Fall 2019 possible.

\mmp{To see if levels can be weighted non-linearly. Can we use the same non-linearity in cost as in distance and does this make sense?}
\mmp{To think whether rounding the $Y^*$ can give us any benefit. I think no, unless we get a better $\delta$, guaranteed. Otherwise we can stay with the exact $\delta$ we have.}

\mmp{Roy and Pokutta have a more recent journal paper published in JMLR, which should be referenced instead of the NiPS paper.}

\end{document}